\documentclass[10pt,twocolumn,letterpaper]{article}

\usepackage{iccv}
\usepackage{times}
\usepackage{epsfig}
\usepackage{graphicx}
\usepackage{amsmath}
\usepackage{amssymb}

\usepackage[utf8]{inputenc}
\usepackage{booktabs}
\usepackage{subfigure}
\usepackage{mathrsfs}
\usepackage{arydshln}
\usepackage[table,xcdraw]{xcolor}
\usepackage{bm}
\usepackage{amsbsy}
\usepackage{multirow}
\usepackage{multicol}
\usepackage{threeparttable}
\usepackage{bbm}
\usepackage[switch]{lineno}

\usepackage[breaklinks=true,bookmarks=false]{hyperref}

\iccvfinalcopy 

\def\iccvPaperID{3037} 

\ificcvfinal\fi

\begin{document}

\title{Instance-wise Hard Negative Example Generation for Contrastive Learning in Unpaired Image-to-Image Translation}


\author{Weilun Wang{\small $~^{1}$}, ~Wengang Zhou{\small $~^{1,2}$}\thanks{Corresponding author: Wengang Zhou and Houqiang Li.}, ~Jianmin Bao{\small $~^{3}$}, ~Dong Chen{\small $~^{3}$}, ~Houqiang Li{\small $~^{1,2\dag}$}\\
\normalsize
$^{1}$ CAS Key Laboratory of GIPAS, EEIS Department, University of Science and Technology of China (USTC) \\
\normalsize
$^{2}$ Institute of Artificial Intelligence, Hefei Comprehensive National Science Center\\
\normalsize
$^{3}$ Microsoft Research Asia\\
\normalsize
{\tt\small wwlustc@mail.ustc.edu.cn, \{zhwg, lihq\}@ustc.edu.cn, \{jianbao, doch\}@microsoft.com}
}

\maketitle
\ificcvfinal\thispagestyle{empty}\fi

\begin{abstract}
\vspace{-1.1em}
Contrastive learning shows great potential in unpaired image-to-image translation, but sometimes the translated results are in poor quality and the contents are not preserved consistently. 
In this paper, we uncover that the negative examples play a critical role in the performance of contrastive learning for image translation.
The negative examples in previous methods are randomly sampled from the patches of different positions in the source image, which are not effective to push the positive examples close to the query examples. 
To address this issue, we present instance-wise hard Negative Example Generation for Contrastive learning in Unpaired image-to-image Translation~(NEGCUT). 
Specifically, we train a generator to produce negative examples online. The generator is novel from two perspectives: 1) it is instance-wise which means that the generated examples are based on the input image, and 2) it can generate hard negative examples since it is trained with an adversarial loss. 
With the generator, the performance of unpaired image-to-image translation is significantly improved. 
Experiments on three benchmark datasets demonstrate that the proposed NEGCUT framework achieves state-of-the-art performance compared to previous methods.
\end{abstract}
\vspace{-1.2em}
\section{Introduction}
\vspace{-0.4em}
Image-to-image translation aims to transfer images from the source domain to the target domain with the content information preserved, which is of significant importance on various applications such as style transfer~\cite{ gatys2016image, johnson2016perceptual, kolkin2019style, luan2017deep}, domain adaption~\cite{bousmalis2017unsupervised, hoffman2018cycada, hoffman2016fcns, tzeng2017adversarial} and image colorization~\cite{zhang2016colorful, zhang2017real, su2020instance, baldassarre2017deep}.
Due to the inconvenience of collecting paired training data, recent methods are usually based on the unpaired setting.
In that case, cycle-consistency loss has been widely used to preserve the consistency between the source images and generated images, for instance, CycleGAN~\cite{zhu2017unpaired}, StarGAN~\cite{choi2018stargan}, UNIT~\cite{liu2017unsupervised} and MUNIT~\cite{huang2018multimodal}.


\begin{figure} 
    \centering 
    \includegraphics[width=\linewidth]{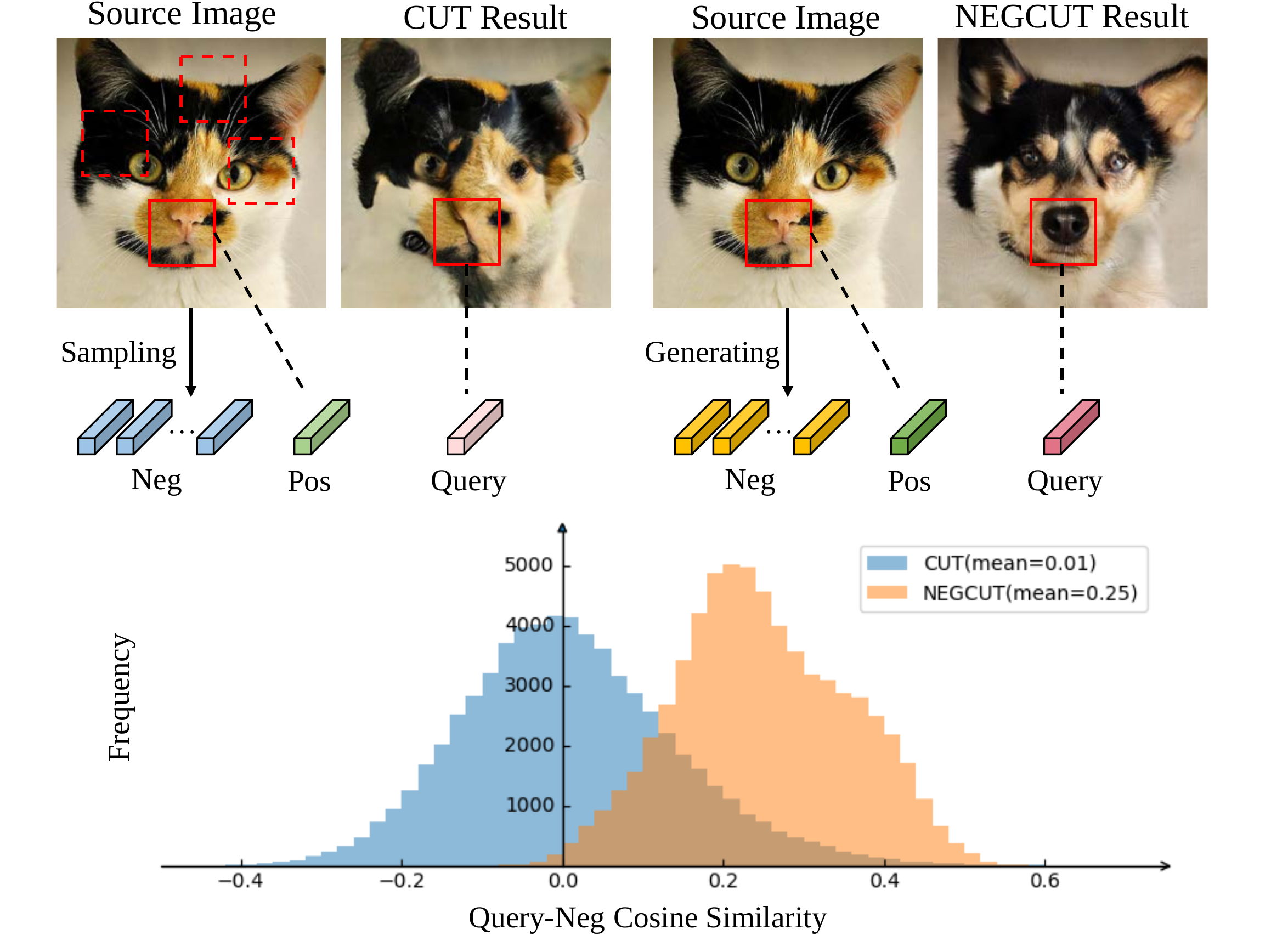}
    \caption{We visualize the generated images along with the distribution of cosine similarity between query and negative samples in CUT~\cite{park2020contrastive} and our method. 
    The blue histogram refers to the distribution in CUT while the orange histogram refers to the distribution in our method. 
    }
    \label{fig:distribution}
    \vspace{-5mm}
\end{figure}

The recently proposed method CUT~\cite{park2020contrastive} introduces contrastive learning in unpaired image-to-image translation and achieves better performance over methods~\cite{DRIT, liu2017unsupervised, zhu2017unpaired} that use cycle-consistency loss. In this paper, we aim to further improve the performance of contrastive learning for unpaired image-to-image translation. We uncover that the performance of contrastive learning relies heavily on the hardness of negative samples. As shown in Figure~\ref{fig:distribution}, 
the negative samples in the method~\cite{park2020contrastive} are randomly sampled from the patches of different positions in the image, which sometimes leads to the translated results in poor quality and the contents not preserved consistently. Also we calculate the cosine similarities between the query patches and negative patches, and we can find that their cosine similarities are around 0. In other words, these negative patches are not
challenging enough to push the positive examples close to the query examples, which will result in the framework not taking full advantage of contrastive learning.

To address the above issue, we present instance-wise hard Negative Example Generation for Contrastive learning in Unpaired image-to-image Translation~(NEGCUT) in this paper.
More precisely, we propose a novel negative generator to excavate hard negative examples.
For a source image, we first extract its features on different layers of image generator encoder and embed them into feature vectors. 
Based on the embedded features from source images, the negative generator produces instance-wise negative examples related to the source image. Moreover, the negative samples should be diverse enough to push the query patch closer to the positive patch. To this end, we add the noise as an extra input for the generator. However, the noise input can probably be ignored for the generator, thus the generator can generate similar examples for different input noises. This is also called the mode collapse issue~\cite{salimans2016improved}. Inspired by the mode seeking loss in MSGAN~\cite{MSGAN}, we introduce diversity loss to the generator to encourage the generator to produce diverse hard negative samples for different input noise.

To generate challenging negative samples for contrastive learning, the main idea is to train the negative generator against the encoder network in an adversarial manner.
Two components in the framework, \emph{i.e.}, the encoder network and negative generator, are updated alternatively to play a min-max game.
On one hand, the encoder network narrows the distance between query and positive samples against hard negative samples to minimize contrastive loss.
On the other hand, the negative generator produces hard negative samples close to the positive samples to maximize contrastive loss.
Intuitively, the framework will reach an equilibrium where the encoder learns detailed and distinguishing representation to discriminate the positive samples from generated hard negative samples.
In Figure~\ref{fig:distribution}, we visualize the generated images along with the distribution of cosine similarity between the query and negative samples in the CUT and NEGCUT.
It is observed that the negative samples produced by negative generator are harder than those sampled in the method~\cite{park2020contrastive}, which push the encoder network to learn distinguishing representation and finally results in fine-grained correspondence of structures and textures. 

Our contributions are summarized as follows,
\begin{itemize}
    \vspace{-1.1em}
    \item We identify that instance-wise negative examples that increase hardness as training process play a critical role in the performance of contrastive learning for unpaired image-to-image translation. 
    \vspace{-1.1em}
    \item We propose a novel framework NEGCUT to mine instance-wise hard negative examples for contrastive learning in unpaired image-to-image translation. 
    \vspace{-1.1em}
    \item Extensive experiments on three benchmark datasets demonstrate the superiority of our method, which achieves new state-of-the-art performance.
    The generated images of our method are of better visual performance with consistent detailed correspondence. 
\end{itemize}

\section{Related Work}

In this section, we briefly introduce the related topics, including contrastive learning, image-to-image translation and hard negative mining. 

\subsection{Image-to-Image Translation}

Image-to-image translation~(I2I)~\cite{lee2018diverse, liu2019few, royer2017xgan, tang2019attentiongan, wang2018high, wu2019transgaga, zhang2019harmonic, zhu2017unpaired, zhu2017toward} aims to transfer images from source to target domain with the content information preserved.
Earlier methods~\cite{wang2018high, isola2017image, chen2017photographic, park2019semantic} apply an adversarial loss~\cite{goodfellow2014generative}, along with a reconstruction loss to train their model based on the paired training data.
However, due to the difficulty of collecting a large amount of paired data, recent methods are usually based on the unpaired setting.
In that cases, cycle-consistency loss has been widely used to preserve the consistency between the source images and generated images instead, for instance, CycleGAN~\cite{zhu2017unpaired}, DiscoGAN~\cite{kim2017learning}, DualGAN~\cite{yi2017dualgan} and U-GAT-IT~\cite{kim2019u}.
Based on the assumption that the generated result should be translated back by an inverse mapping, cycle-consistency learns the mapping from target to source domain and check whether the source images are reconstructed.
However, the assumption is overly strict compared to the actual situation, where the images between the two domains are not one-to-one mapping.
In view of this, CUT~\cite{park2020contrastive} involves contrastive learning in unpaired image-to-image translation to learn the correspondence between source and generated images, which outperforms previous methods using cycle-consistency loss.

\begin{figure*}[t]
	\centering
	\includegraphics[width=\linewidth]{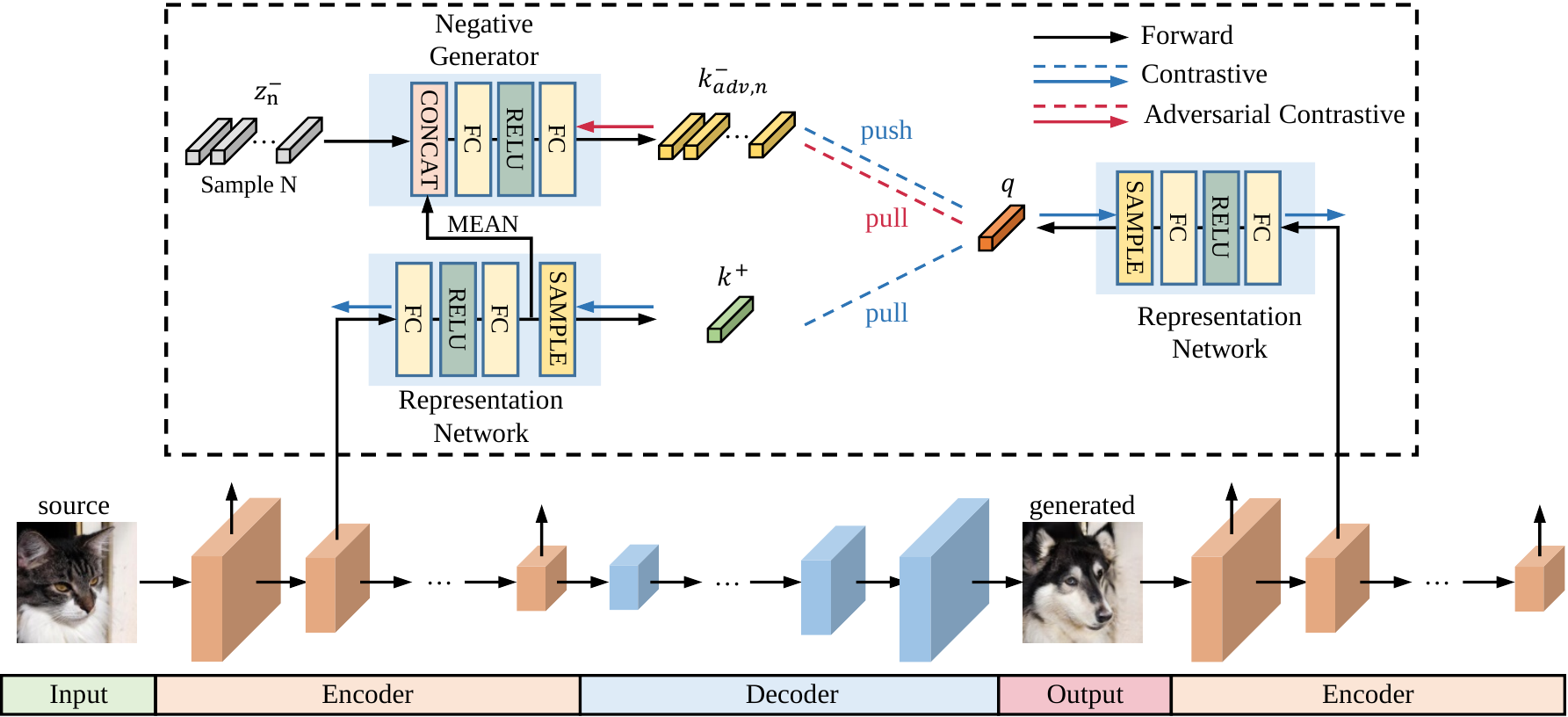}
	\caption{The overview of our NEGCUT framework.
	We perform hard negative example generation for adversarial contrastive learning on multiple layers of the image generator encoder.
	The black arrows show the forward propagation of our framework while the blue and red arrows show the backward propagation of contrastive loss and adversarial contrastive loss, respectively.
	On each layer, the representation network randomly samples the source and translated features at the spatial dimension, and produces the query and positive samples.
	The negative generator produces challenging negative samples by the mean vector of features from the representation network.
	The query, positive and generated negative samples are involved for contrastive learning in an adversarial manner.}
	\vspace{-4mm}
	\label{fig:framework}
\end{figure*}

\subsection{Contrastive Learning}

Contrastive learning is a framework that learns representation by comparing similar and dissimilar pairs.
Recent methods~\cite{bachman2019learning, chen2020simple, he2020momentum, henaff2020data, hjelm2018learning, oord2018representation} based on the theory of maximizing mutual information have achieved wide success on unsupervised representation learning.
These methods take full advantage of noise-contrastive estimation~\cite{gutmann2010noise}, mapping the images into an embedding space where associated samples are brought together in contrast with unrelated samples.
For a single query sample, the associated samples are referred to as positive samples while the unrelated samples are referred to as negative samples.
With similarity measured by dot production, a form of a contrastive loss, called InfoNCE, is proposed as a representative loss function for noise-contrastive estimation.

\subsection{Hard Example Mining}
Hard example mining is a classic method to solve the problem of sample imbalance in several areas, \emph{i.e.}, object detection and unsupervised representation learning. 
In earlier methods, hard example mining are used to optimize SVMs~\cite{felzenszwalb2009object}, shallow neural networks~\cite{rowley1998neural} and boosted decision trees~\cite{dollar2009integral}.
Recent work~\cite{loshchilov2015online, simo2014fracking, wang2015unsupervised, shrivastava2016training, lin2017focal, hu2020adco, kalantidis2020hard} selects hard examples for training deep networks.
In \cite{simo2014fracking}, an image descriptor is learned to independently select the hard positive and negative samples from a large set.
In \cite{loshchilov2015online} and \cite{shrivastava2016training}, online hard examples selection is investigated on image classification and object detection, respectively.
Lin \emph{et~.al}~design a novel focal loss \cite{lin2017focal} to focus training on a sparse set of hard examples, which addresses the imbalance between different classes in object detection.
In unsupervised representation learning, a triple loss is used~\cite{wang2015unsupervised} to mine the hard negative samples from a large set.
In \cite{hu2020adco}, adversarial learning is involved to generate challenging negative samples for unsupervised representation learning.

\vspace{-0.5em}
\section{Methods}



    
    

In this paper, we present a novel framework NEGCUT to mine instance-wise hard negative examples for contrastive learning in unpaired image-to-image translation.
Different from previous work which randomly samples negative examples from the patches in the image, our method generates instance-wise hard negative examples through adversarial learning.
With the produced hard negative examples, our framework can generate images with detailed and fine-grained correspondence on structures and textures.
The rest of this section is organized as follows:
We begin with reviewing the related method in previous work in Sec.~\ref{sec:review}.
In Sec.~\ref{sec:negcut}, we outline the NEGCUT framework and introduce the details of hard negative example generation through adversarial learning.
Finally, we discuss the objective function utilized in our framework in Sec.~\ref{sec:loss}.

\subsection{Preliminaries and Motivation}
\label{sec:review}
We first briefly review the method leveraging contrastive learning in unpaired image-to-image translation developed in CUT~\cite{park2020contrastive}.
To generate images of target domain with the content information maintained, the main idea is to learn the correspondence between the source and generated images.
Compared with previous methods using cycle-consistency loss, CUT applies contrastive loss to learn the correspondence instead, which directly maximizes the mutual information between the source and generated images.
The contrastive loss is formulated as follows,
\begin{equation}
    \begin{split}
            l(\mathbf{q},~&\mathbf{k}^+,~\mathbf{k}^-) = \\
            &-\mathrm{log}[\frac{\mathrm{exp}(\mathbf{q} \cdot \mathbf{k}^+ / \tau)}{\mathrm{exp}(\mathbf{q} \cdot \mathbf{k}^+/ \tau) + \sum_{n=1}^N\mathrm{exp}(\mathbf{q} \cdot \mathbf{k}_n^- / \tau)}],
    \end{split}
\end{equation}
where $\mathbf{q}$ is the query samples from the generated image, $\mathbf{k}^+$ is the positive samples from the corresponding position of the query in the source image, $\mathbf{k}_n^-$ is the negative samples from the other positions in the source images, and $\tau$ is the temperature factor.

CUT develops the contrastive learning in a multi-layer patch-wise manner, which is formulated as follows,
\begin{equation}
    \mathcal{L}_{\mathrm{PatchNCE}}(G, H, \mathbf{X}) = \mathbb{E}_{x \sim X} \sum_{l=1}^L \sum_{s=1}^{S_l} l(\mathbf{q}_{l,s}, \mathbf{k}_{l,s}^+, \mathbf{k}_{l,s}^-),
\end{equation}
where $\mathbf{q}_{l,s}$, $\mathbf{k}_{l,s}^+$ and $\mathbf{k}_{l,s}^-$ are extracted from the features of source image $\mathbf{X}$ and generated image $\mathbf{Y}$ at different intermediate layers $l$ of generator encoder.
With the contrastive loss, the generator learns to narrow the distance between the query and positive samples against negative samples at different layers, which is equivalent to maximizing the mutual information between the source and generated images.

By replacing the cycle-consistency loss with the contrastive loss, CUT generates more realistic and corresponding images compared with previous methods.
However, the randomly-sampled negative examples in CUT cannot take full advantage of contrastive learning.
The approach to estimating the negative examples plays a critical role in the performance of contrastive learning.
Negative examples in CUT are not challenging enough to push the encoder network to learn distinguishing representation, which leads to the translated results in poor quality and the contents not preserved consistently.
Different from these, we propose a novel framework NEGCUT to mine instance-wise hard negative samples for unpaired image-to-image translation through adversarial learning.

\subsection{NEGCUT}
\label{sec:negcut}
\subsubsection{Framework Architecture}
\label{sec:framework}
\hspace{1em} Figure~\ref{fig:framework} gives an overview of our framework, which consists of \emph{Image Generator}, \emph{Representation Network} and \emph{Negative Generator}.
Image generator $G$ takes the source image $\mathbf{X}$ as input and generates the translated image $\mathbf{Y}$.
Regarding two variants of a single image, \emph{i.e.}, the source image $\mathbf{X}$ and the generated image $\mathbf{Y}$, we conduct multi-layer patch-wise contrastive learning to learn the correspondence between these two images.
On a certain layer of the image generator encoder, the query and positive samples are produced by the representation network through embedding the spatially sampled feature vectors into high-dimensional representation space.

To increase the similarity between the query and positive samples, the negative generator mines instance-wise hard negative samples against positive samples.
Based on the embedded features of the source image, diverse challenging negative examples are generated by taking various randomly-sampled noise vectors as input.  
In our framework, the encoder network~(\emph{i.e.}, the image generator and representation network) and the negative generator are alternately updated with the adversarial contrastive loss.
With more challenging negative examples produced by the negative generator, the encoder network will learn distinguishing representation to discriminate positive samples from the challenging negative samples, which leads to fine-grained and robust correspondence between the source and generated images. 
Additionally, a discriminator is applied to ensure the domain and realness of the generated image.


\subsubsection{Hard Negative Example Generation}
\label{sec:neg-gen}

\hspace{1em} In this section, we formally present hard negative example generation for contrastive learning in unpaired image-to-image translation.
As shown in Figure~\ref{fig:framework}, we perform contrastive learning on multiple layers of the image generator encoder.
For a certain layer, we employ a representation network $H^i(\cdot)$ to embed the feature of different patches.
The representation network is a 2-layer MLP network independently mapping the feature vector at each pixel from the source and translated images to a $M$-dimension vector.
Based on the feature after mapping, we randomly sample $S$ positions in the spatial dimension and take the normed vectors as query and positive samples for contrastive learning, which is formulated as follow,
\begin{equation}
    q = \frac{H_s^i(\mathbf{F}^\mathbf{Y}_i)}{\|H_s^i(\mathbf{F}^\mathbf{Y}_i)\|_2},
    k^+ = \frac{H_s^i(\mathbf{F}^\mathbf{X}_i)}{\|H_s^i(\mathbf{F}^\mathbf{X}_i)\|_2},
\end{equation}
where $\mathbf{F}^\mathbf{X}_i$ and $\mathbf{F}^\mathbf{Y}_i$ are the source features and the translated features at the $i$-th layer of image generator encoder, respectively.
$H_s^i(\mathbf{F}^\mathbf{X}_i)$ and $H_s^i(\mathbf{F}^\mathbf{Y}_i)$ refers to the $s$-th positive and query examples sampled, respectively.

To push the positive sample close to the query sample, we generate challenging negative samples with a carefully designed multi-layer negative generator $\{N^0, N^1, \cdots, N^l \}$.
Base on the spatially-average features from representation network $\overline{H^i(\mathbf{F}^\mathbf{X}_i)}$, the negative generator produces hard negative samples with noise vector $z_n$, which is formulated as follows,
\begin{equation}
    k^-_{\mathrm{adv},n} = \frac{N^i(\overline{H^i(\mathbf{F}^\mathbf{X}_i)}; z_n)}{\|N^i(\overline{H^i(\mathbf{F}^\mathbf{X}_i)}; z_n)\|_2}.
\end{equation}
For a positive sample, we generate multiple negative examples through sampling various noise vectors from standard Gaussian distribution.

To generate challenging negative samples for contrastive learning, the main idea is to train the negative generator against the encoder network in an adversarial manner, which is formulated as follows,
\begin{equation}
    \begin{split}
    &\min_{\theta_\mathcal{H}, \theta_G} \max_{\theta_\mathcal{N}} ~ l(\mathbf{q}, \mathbf{k}^+, \mathbf{k}_{adv}^-) = \\
     &-\mathrm{log}[\frac{\mathrm{exp}(\mathbf{q} \cdot \mathbf{k}^+ / \tau)}{\mathrm{exp}(\mathbf{q} \cdot \mathbf{k}^+/ \tau) + \sum_{n=1}^N \mathrm{exp}(\mathbf{q} \cdot \mathbf{k}_{\mathrm{adv},n}^- / \tau)}].
    \end{split}
    \label{equ:adv_loss}
\end{equation}
From Equation~\eqref{equ:adv_loss}, it is observed that the encoder network~(\emph{i.e.}, the representation network $\mathcal{H} = \{H^0, H^1, \cdots, H^l \}$ and the image generator $G$) narrows the distance between the query samples and positive samples against the negative samples to minimize contrastive loss.
On the contrary, the negative generator $\mathcal{N} = \{N^0, N^1, \cdots, N^l \}$ produces challenging negative examples to maximize the contrastive loss.
Intuitively, the encoder network and the negative generator will reach an equilibrium by alternate training, where the negative generator produces challenging negative samples and the encoder network learns distinguishing representation to discern the positive samples from the negative samples.

In Figure~\ref{fig:framework}, we further illustrate how the negative generator, representation network and image generator are updated.
The negative generator is first updated with negative contrastive loss, which is formulated as follows,
\begin{equation}
    \theta_{N^i} \leftarrow \theta_{N^i} + \eta_\mathcal{N} \frac{\partial l(\mathbf{q}, \mathbf{k}^+, \mathbf{k}^-_{adv})}{\partial \theta_{N^i}}.
\end{equation}
The backpropagation of the negative contrastive loss is cut off before the representation network and does not affect the weights of the representation network and image generator.
After that, the representation network is updated with positive contrastive loss, which is formulated as follow,
\begin{equation}
    \theta_{H^i} \leftarrow \theta_{H^i} - \eta_\mathcal{H} \frac{\partial l(\mathbf{q}, \mathbf{k}^+, \mathbf{k}^-_{adv})}{\partial \theta_{H^i}}.
\end{equation}
Since the contrastive learning is developed in a multi-layer manner, the total adversarial contrastive loss for the negative generator and representation network is formulated as follows,
\begin{equation}
    \mathcal{L}_{AdCont} = \mathbb{E}_{x \sim X} \sum_{l=1}^L \sum_{s=1}^{S_l} l(\mathbf{q}_{l,s}, \mathbf{k}_{l,s}^+, \mathbf{k}_{adv, l ,s}^-).
\end{equation}

The image generator is trained along with the representation network.
Through the back-propagation of adversarial contrastive loss, the image generator receives the gradient at different layers of the encoder.
The image generator is updated with the summation of these gradient, which is formulated as follows,
\begin{equation}
    \begin{split}
    \theta_G \leftarrow \theta_G - \eta_G \sum_{i=0}^l (& \frac{\partial l(\mathbf{q}, \mathbf{k}^+, \mathbf{k}^-_{adv})}{\partial \mathbf{F}^\mathbf{X}_i} \frac{\partial \mathbf{F}^\mathbf{X}_i}{\partial \theta_G} \\
    &+ \frac{\partial l(\mathbf{q}, \mathbf{k}^+, \mathbf{k}^-_{adv})}{\partial \mathbf{F}^\mathbf{Y}_i} \frac{\partial \mathbf{F}^\mathbf{Y}_i}{\partial \theta_G}),
    \end{split}
\end{equation}
where $\mathbf{F}^\mathbf{X}_i$ and $\mathbf{F}^\mathbf{Y}_i$ are the features of the source and translated images at the $i$-th layer of encoder, respectively.

However, when the adversarial contrastive loss is the only function used to update the negative generator, it is observed that the generated negative examples lose diversity and collapse to one negative example.
This is because the adversarial contrastive loss focuses on generating hard negative samples rather than diverse negative samples, though diversity is helpful for the performance. 
To this end, we introduce the diversity loss to generate diverse challenging negative samples with different input noise.
The diversity loss encourages the generation of distinctive results when different noise vectors are brought in, which is formulated as follows,
\begin{equation}
    \mathcal{L}_{div} = - \|N^i(\overline{H^i(\mathbf{X}_i)}, z_1) - N^i(\overline{H^i(\mathbf{X}_i)}, z_2) \|_1,
\end{equation}
where $z_1$ and $z_2$ are two different input noise randomly sampled from standard Gaussian distribution.


\subsection{Other Objectives}
\label{sec:loss}
Besides the adversarial contrastive loss and diversity loss mentioned above, our framework is also optimized by generative adversarial loss.

\noindent \textbf{Generative Adversarial Loss.}
Since the ground-truth images are unavailable in unpaired image-to-image translation, we develop adversarial learning to constrain the realness and the domain of the generated images.
For the image generator $G(\cdot)$ and the discriminator $D(\cdot)$, we unitize the $\mathrm{LSGAN}_{100}$ loss~\cite{mao2017least}, which is formulated as follows,
\begin{equation}
\begin{split}
    \mathcal{L}_{gan}^D &= \mathbb{E}_{x_r}[(1-\mathbf{D}(x_r))^2] + \mathbb{E}_{x_f}[\mathbf{D}(x_f)^2], \\
    \mathcal{L}_{gan}^G &= \mathbb{E}_{x_f}[(1-\mathbf{D}(x_f)^2],
\end{split}
\end{equation}
where $x_r$ and $x_f$ indicates the real image distribution and the generated images distribution, respectively.

\noindent \textbf{Overall Loss.}
The overall loss for the negative generator and encoder network is the weighted summation of above losses, which is formulated as follows,
\begin{equation}
    \begin{split}
            \mathcal{L}_\mathcal{H} &= \mathcal{L}_{AdCont}, \\
            \mathcal{L}_{G} &= \mathcal{L}_{AdCont} + \lambda_1 \mathcal{L}_{gan}^{G}, \\
            \mathcal{L}_\mathcal{N} &= -\mathcal{L}_{AdCont} + \lambda_2 \mathcal{L}_{div}, \\
    \end{split}
\end{equation}
where $\lambda_1$ and $\lambda_2$ are the trade-off parameters balancing different losses. 
In our experiments, $\lambda_1$ and $\lambda_2$ are set to 1 and 1, respectively.

\section{Experiment}

\begin{figure*}[t]
	\centering
	\includegraphics[width=\linewidth]{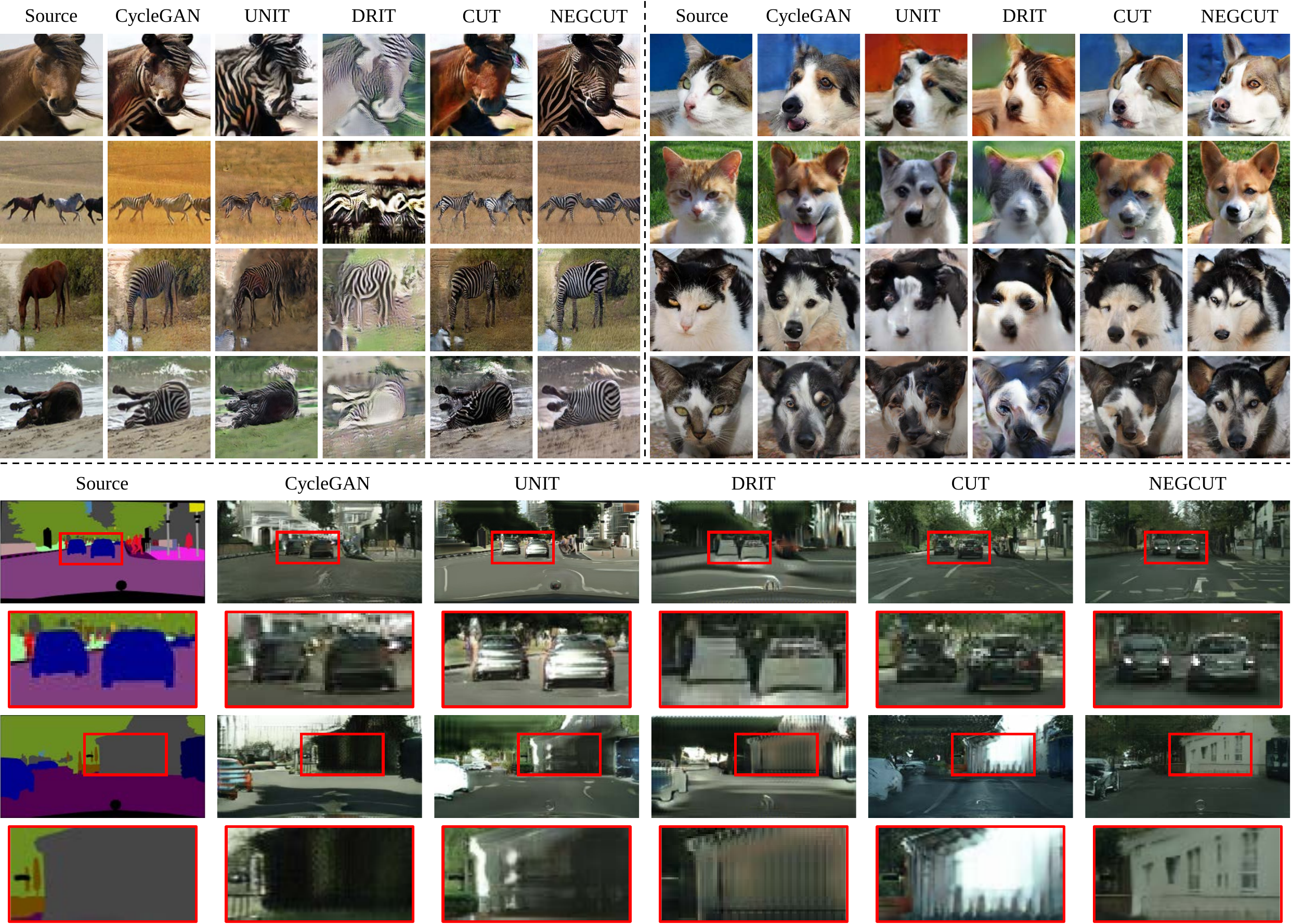}
	\vspace{1mm}
	\caption{Qualitative results with the four challenging methods, \emph{i.e.}, CycleGAN~\cite{zhu2017unpaired}, UNIT~\cite{liu2017unsupervised} , DRIT~\cite{DRIT}, CUT~\cite{park2020contrastive}, on three benchmark datasets.
	Compared with previous methods, the generated images of our method show superior performance with correct correspondence between the source and generated image.}
	\vspace{-3mm}
	\label{fig:compare}
\end{figure*}

\subsection{Experiment Setup}
\noindent \textbf{Datasets.} 
To demonstrate the superiority of our method, we train and test our method on three benchmark datasets, \emph{i.e.}, \emph{Cityscapes}~\cite{cordts2016cityscapes}, \emph{Cat$\rightarrow$Dog}~\cite{choi2020starganv2} and \emph{Horse$\rightarrow$Zebra}~\cite{zhu2017unpaired} datasets with translation between various different domains.
The \emph{Cityscapes} dataset contains a diverse set of images recorded in the street scenes with high-quality pixel-level annotations.
The \emph{Cat$\rightarrow$Dog} dataset is a dataset of 10,000 high-quality cat and dog face images extracted from the \emph{AFHQ} dataset.
The \emph{Horse$\rightarrow$Zebra} dataset consists of about 2,500 images of horse and zebra in different scenes.
We learn the translation from semantic masks to real images, from cat images to dog images and from horse images to zebra images on three datasets, respectively.
For all the datasets, we resize images to the same resolution of 256 $\times$ 256 to train our network.

\noindent \textbf{Implementation Details.}
To make a fair comparison, we set the hyperparameters consistent with previous methods~\cite{park2020contrastive}.
We conduct our adversarial contrastive learning on the $1$-st, $5$-th, $9$-th, $13$-th, $17$-th layers of the generator encoder.
The number of negative samples for contrastive learning is set to 256 in our framework.
The dimension of the query, positive and negative samples is set to 256.
For the whole framework, we utilize Adam optimizer~\cite{kingma2014adam}.
The training lasts 400 epochs in total.
The learning rate is set to 2e-4 and linearly reduces after 200 epochs.
The whole framework is implemented by Pytorch and we perform experiments on NVIDIA RTX 3090Ti.

\noindent \textbf{Evaluation Metrics.}
We evaluate the realness of generated images by the FID metric.
FID measures the distance between two sets of images. 
To calculate the FID metric, we first embed the generated images and ground-truth images into the feature space with an Inception model~\cite{szegedy2016rethinking}. 
The FID metric is computed by the mean value and covariance of the generated image set $(\mu_\mathbf{Y}, \Sigma_\mathbf{Y})$ and the ground-truth image set $(\mu_\mathbf{\hat{Y}}, \Sigma_\mathbf{\hat{Y}})$:
\begin{equation}
    \mathrm{FID}(\mathbf{Y}, \mathbf{\hat{Y}}) = \| \mu_\mathbf{Y} - \mu_\mathbf{\hat{Y}} \|_2^2 + \mathrm{Tr}(\Sigma_\mathbf{Y} + \Sigma_\mathbf{\hat{Y}} - 2(\Sigma_\mathbf{Y}\Sigma_\mathbf{\hat{Y}})^{\frac{1}{2}}).
\end{equation}
In addition, to evaluate the relevance between source images and generated images, we apply several metrics different from FID on the \emph{Cityscapes} dataset.
With a pre-trained segmentation model~\cite{Yu2017}, we calculate the mAP, pixel accuracy~(pAcc) and class accuracy~(cAcc) metrics on the source semantic labels and generated real images.
The higher mAP, pAcc and cAcc represent that the generated images are more relevant to source semantic labels.

\begin{figure}[t]
	\centering
	\includegraphics[width=\linewidth]{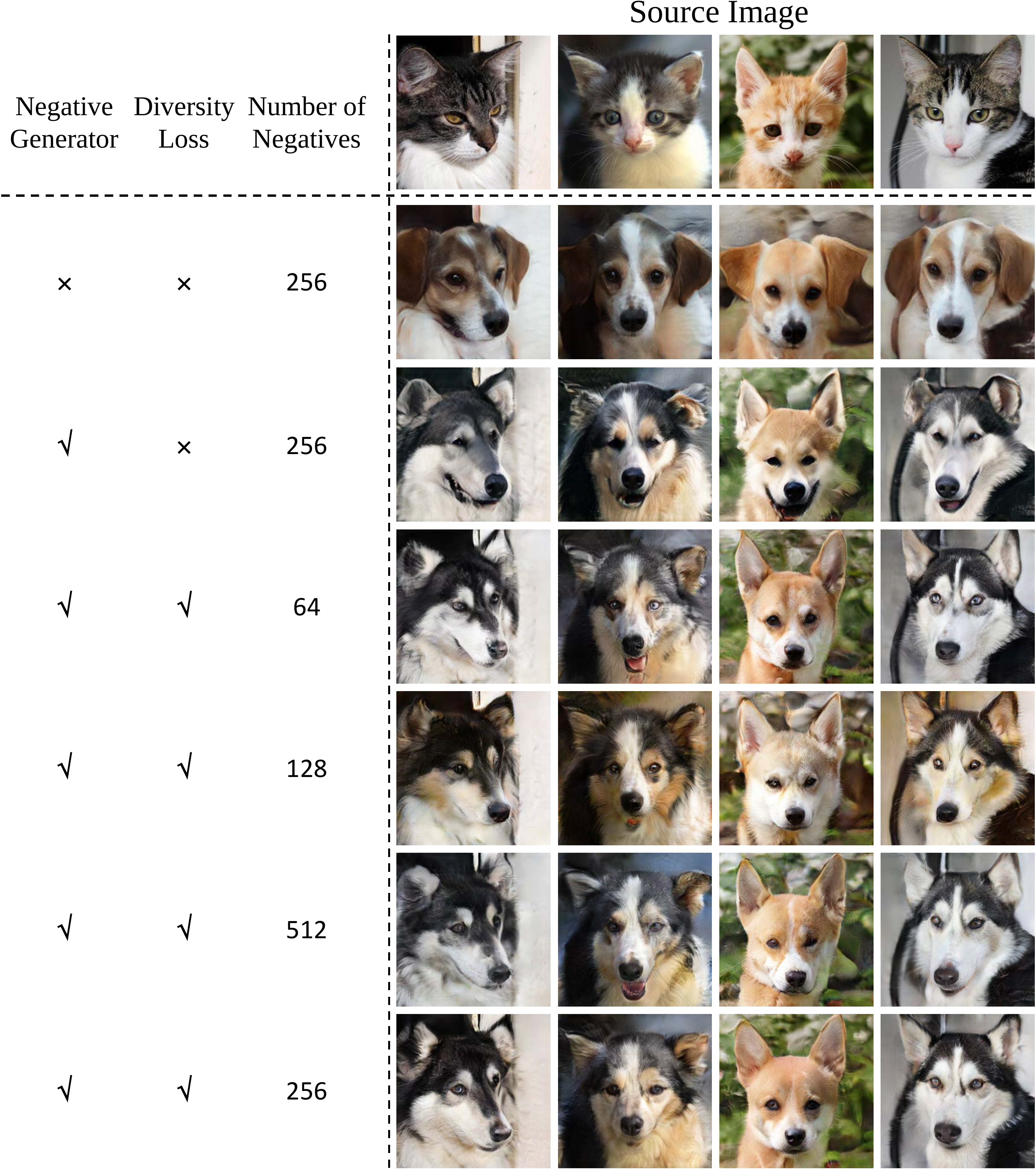}
	\vspace{1mm}
	\caption{Qualitative comparison among different designs of negative generator.
    When the negative generator and diversity loss are employed and the number of negatives is set to 256, the generated images have the best visual quality and most correct correspondence.}
	\vspace{-4mm}
	\label{fig:ablation}
\end{figure}


\subsection{Comparison with the State-of-the-art Methods}
We compare our method with several state-of-the-art methods of unpaired image-to-image translation, \emph{i.e.}, CUT~\cite{park2020contrastive}, CycleGAN~\cite{zhu2017unpaired} and DRIT~\cite{DRIT}.
The quantitative result on three benchmark datasets is shown in Table~\ref{tab:comparison}.
From the table, it is observed that our method achieves new state-of-the-art performance on three datasets.
Compared with the most challenging method, \emph{i.e.} CUT, our method outperforms it $14.0\%$, $26.6\%$ and $13.0\%$ relatively on FID metric on three datasets.
Additionally, since only the image generator is used at inference time, NEGCUT does not introduce extra test time consumption compared with CUT.

Furthermore, we make a qualitative evaluation on three datasets with several competitive methods, \emph{i.e.} CUT, CycleGAN, UNIT and DRIT.
From Figure~\ref{fig:compare}, it is observed that the images generated by our method have better visual performance compared with previous methods.
Especially, our generated images keep a better correspondence with the source images compared with the most challenging method CUT.
This benefits from the challenging negative samples generated by the negative generator.
The negative examples sampled randomly in CUT help the network learn the correspondence between source images and generated images at the beginning, but become less and less effective as the training process proceeds.
In contrast, our negative samples produced by negative generator keep challenging via adversarial learning, which forces the image generator and representation network to learn the fine-grained correspondence.
Due to this reason, the images generated by our method have better correspondence with the source images in details, \emph{i.e.}, textures and postures.

\begin{table}[t]
    \scriptsize
    \centering
    \resizebox{0.48\textwidth}{!}{
        \begin{tabular}{l @{\hskip 4mm} c@{\hskip 4mm}ccc @{\hskip 4mm} c @{\hskip 4mm} c}
        \toprule
        \multirow{2}{*}{\textbf{Method}}  & \multicolumn{4}{c}{\textbf{Cityscapes}} & {\textbf{Cat$\rightarrow$Dog}} & {\textbf{ H$\rightarrow$Z}} \\ 
        \cmidrule(r){2-5} \cmidrule(r){6-6} \cmidrule(r){7-7}
        & {\bf mAP}$\uparrow$ & {\bf pAcc}$\uparrow$ & {\bf cAcc}$\uparrow$ & {\bf FID}$\downarrow$ & {\bf FID}$\downarrow$ & {\bf FID}$\downarrow$\\
            \midrule
        {CycleGAN~\cite{zhu2017unpaired}} & 20.4 & 55.9 & 25.4 & 76.3 & 85.9 & 77.2\\ 
        {UNIT~\cite{liu2017unsupervised}} & 16.9 & 56.5 & 22.5 & 91.4 & 104.4 & 133.8\\
        {DRIT~\cite{DRIT}} & 17.0 & 58.7 & 22.2 & 155.3 & 123.4 & 140.0\\ \cdashline{1-7}
        {Distance~\cite{Benaim2017OneSidedUD}} & 8.4 & 42.2 & 12.6 & 81.8 & 155.3 & 72.0 \\ 
        {SelfDistance~\cite{Benaim2017OneSidedUD}} & 15.3 & 56.9 & 20.6 & 78.8 & 144.4 & 80.8\\ 
        {GCGAN~\cite{FuCVPR19-GcGAN}} & 21.2 & 63.2 & 26.6 & 105.2 & 96.6 & 86.7\\ \cdashline{1-7}
        {CUT~\cite{park2020contrastive}} & 24.7 & 68.8 & 30.7 & 56.4 & 76.2 & 45.5\\
        {FastCUT~\cite{park2020contrastive}} & 19.1 & 59.9 & 24.3 & 68.8 & 94.0 & 73.4\\
        {NEGCUT}  & {\bf 27.6} & {\bf 71.4} & {\bf 35.0} & {\bf 48.5} & {\bf 55.9} & {\bf 39.6}\\
        \bottomrule
        \end{tabular}
    }
    \vspace{0.5mm}
    \caption{Comparison with state-of-the-art methods on unpaired image translation, \emph{i.e.} CycleGAN, UNIT, DRIT, CUT, \emph{etc.}  
    H$\rightarrow$Z refers to the \emph{Horse$\rightarrow$Zebra} dataset.
    $\uparrow$ indicates the higher the better, while $\downarrow$ indicates the lower the better.
    It is notable that our method outperforms previous methods on various metrics. 
    }
    \vspace{-3mm}
    \label{tab:comparison}
\end{table}

\begin{table}[t]
    \small
    \centering
    \resizebox{0.48\textwidth}{!}{
        \begin{tabular}{c@{\hskip 1mm}c@{\hskip 1mm}c @{\hskip 2mm} c@{\hskip 2mm}ccc @{\hskip 2mm} c @{\hskip 2mm} c}
        \toprule
        \multicolumn{3}{c}{\textbf{Settings}}  & \multicolumn{4}{c}{\textbf{Cityscapes}} & {\textbf{Cat$\rightarrow$Dog}} & {\textbf{ H$\rightarrow$Z}} \\ 
        \cmidrule(r){1-3}\cmidrule(r){4-7} \cmidrule(r){8-8} \cmidrule(r){9-9}
        {\bf Negative} & {\bf Diversity} & {\bf Number} & \multirow{2}{*}{\bf mAP$\uparrow$} & \multirow{2}{*}{\bf pAcc$\uparrow$} & \multirow{2}{*}{\bf cAcc$\uparrow$} & \multirow{2}{*}{\bf FID$\downarrow$} & \multirow{2}{*}{\bf FID$\downarrow$} & \multirow{2}{*}{\bf FID$\downarrow$}\\
        {\bf Generator} & {\bf Loss} & {\bf of Neg.} & & &  &  &  & \\ 
            \midrule
        $\times$ & $\times$ & 256 & 27.3 & {\bf 71.9} & 34.5 & 49.7 & 110.9 & 59.6\\ 
        $\checkmark$ & $\times$ & 256 & 27.0 & 71.1 & 33.7 & 91.5 & 83.0 & 72.1 \\ \cdashline{1-9}
        $\checkmark$ & $\checkmark$ & 64 & 26.9 & 71.1 & 33.7 & 49.7 & 59.3 & 59.3\\ 
        $\checkmark$ & $\checkmark$ & 128 & 27.2 & 71.4 & 33.9 & 49.8 & 86.9 & 51.8\\ 
        $\checkmark$ & $\checkmark$ & 512 & 27.3 & 71.3 & 34.3 & 51.2 & 62.8 & 44.0\\ \cdashline{1-9}
        $\checkmark$ & $\checkmark$ & 256 & {\bf 27.6} & 71.4 & {\bf 35.0} & {\bf 48.5} & {\bf 55.8} & {\bf 39.6}\\
        \bottomrule
        \end{tabular}
    }
    \vspace{0.5mm}
    \caption{Ablation study for several different designs, \emph{i.e.}, negative generator, diversity loss and the number of negative samples.
    H$\rightarrow$Z refers to the \emph{Horse$\rightarrow$Zebra} dataset.
    $\uparrow$ indicates the higher the better, while $\downarrow$ indicates the lower the better.
    Without negative generator or the diversity loss, the produced negative examples are not challenging enough, which leads to inferior performance under most of the indicators on three datasets.
    }
    \vspace{-4mm}
    \label{tab:ablation}
\end{table}

\subsection{Ablation Study}

We perform several ablation experiments to verify the effectiveness of several designs in our framework, \emph{i.e.} negative generator, diversity loss and number of negative samples.
We report the quantitative results in Table~\ref{tab:ablation}.


To evaluate the necessity of generating instance-wise negative samples with negative generator, we design a variant without negative generator.
As an alternative, we directly update the negative samples in the feature vector space in this variant.
In such case, the learned negative samples are widely distributed in the feature space and unrelated to the source instance.
From Table~\ref{tab:ablation}, it is observed that, without the negative generator, the framework obtains an inferior performance under most of the indicators, which verifies the effectiveness of generating \emph{instance-wise} negative samples.
After that, we conduct an ablation study on the diversity loss by comparing the framework with and without the diversity loss.
In Table~\ref{tab:ablation}, it demonstrates that the framework with diversity loss outperforms that without diversity loss. 
This is because, in the variant without diversity loss, the produced negative samples lose diversity in the early stage of training and maintain less diverse during training. 
Under this situation, the negative generator fails to produce challenging negative samples, which leads to the poor performance.
Additionally, we analyze the visual results under these different settings.
In Figure~\ref{fig:ablation}, it can be seen that, without the negative generator or diversity loss, the generated image has a far inferior visual quality on fidelity and correspondence between the source and generated image.

Based on the framework with negative generator and diversity loss, we further make an ablation study on the number of negative samples. 
From Table~\ref{tab:ablation}, it can be seen that the performance reaches the top when the number of negative samples equal to 256.
When the number of negative samples is more than 256, the main challenging negative samples have been contained in the 256 negative samples.
The extra generated negative samples may contain some irrelevant disturbance resulting in inferior performance, and increase the computation cost notably.
Nevertheless, too few negative samples may result in the ineffectiveness to push positive samples closer to the query.
Comprehensively considering the performance and computational consumption, the best number of negative samples is 256 in our framework.
Furthermore, we compare the generated images under different numbers of negative examples.
In Figure~\ref{fig:ablation}, it is observed that, when the number of negative samples is set to 256, the generated images have the best visual quality and most correct correspondence.

\subsection{Visualization of Hard Negative Examples.}
\begin{figure}[t]
	\centering
	\includegraphics[width=\linewidth]{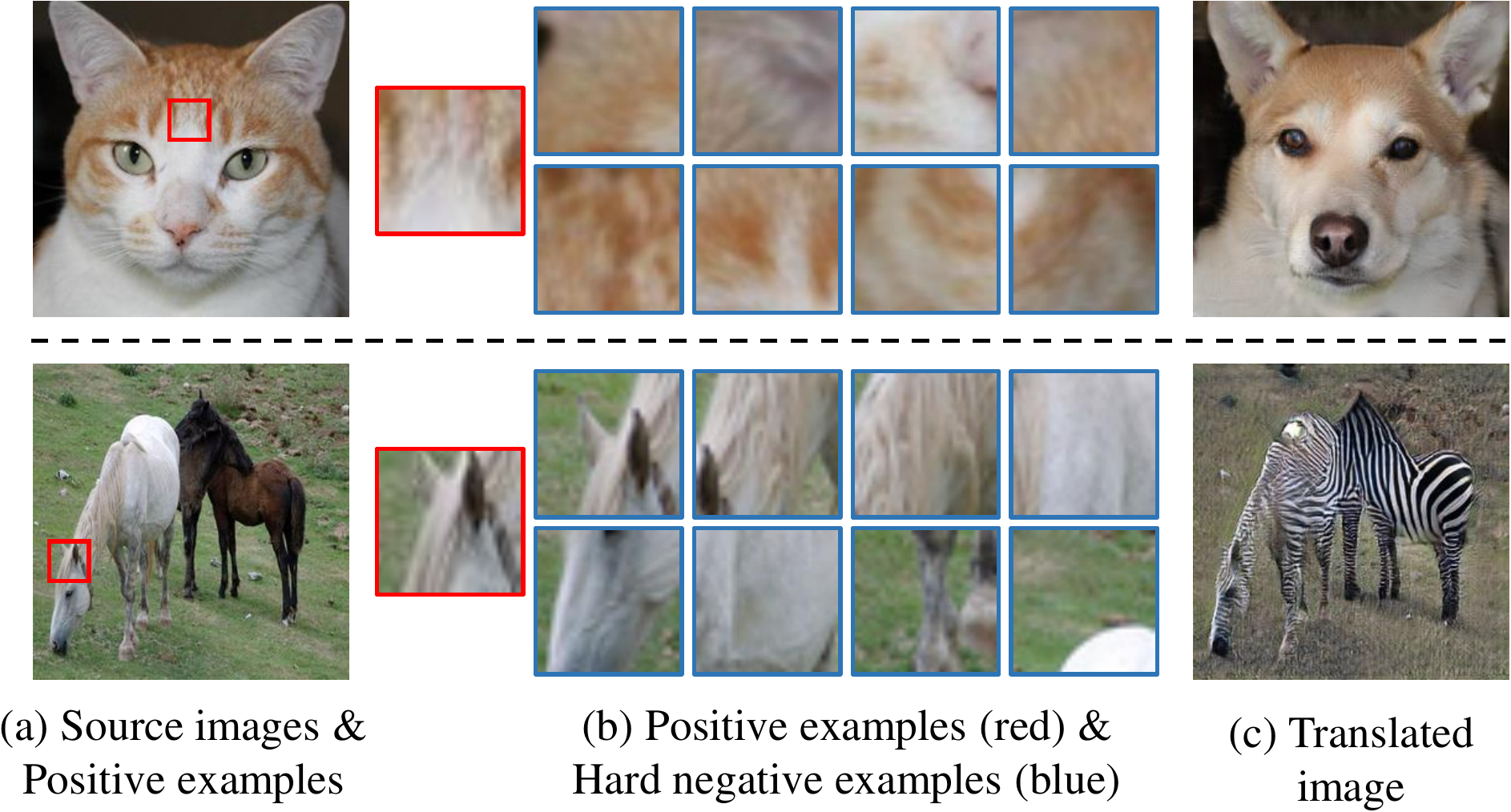}
	\caption{
	Visualization of negative examples by retrieving regions based on generated features.
	We visualize 8 hard negative examples by retrieving the most related patches in the image.
	It is observed that the retrieved patches share similar semantic meanings with the query patch in structure and texture.
	}
	\vspace{-3mm}
	\label{fig:retrieve}
\end{figure}

To further demonstrate the effect of hard negative examples, we visualize hard negative examples by retrieving regions based on generated features.
In Figure~\ref{fig:retrieve}, we first retrieve 8 hard negative examples based on each query feature.
After that, we visualize these hard negative examples by retrieving the most related patches in the image. 
It is observe that the retrieved hard negative examples share similar semantic meanings with the query patch in structure and texture. 
This indicates that the generated hard examples can encourage the model to generate content consistent results.

\begin{figure}[t]
	\centering
	\includegraphics[width=\linewidth]{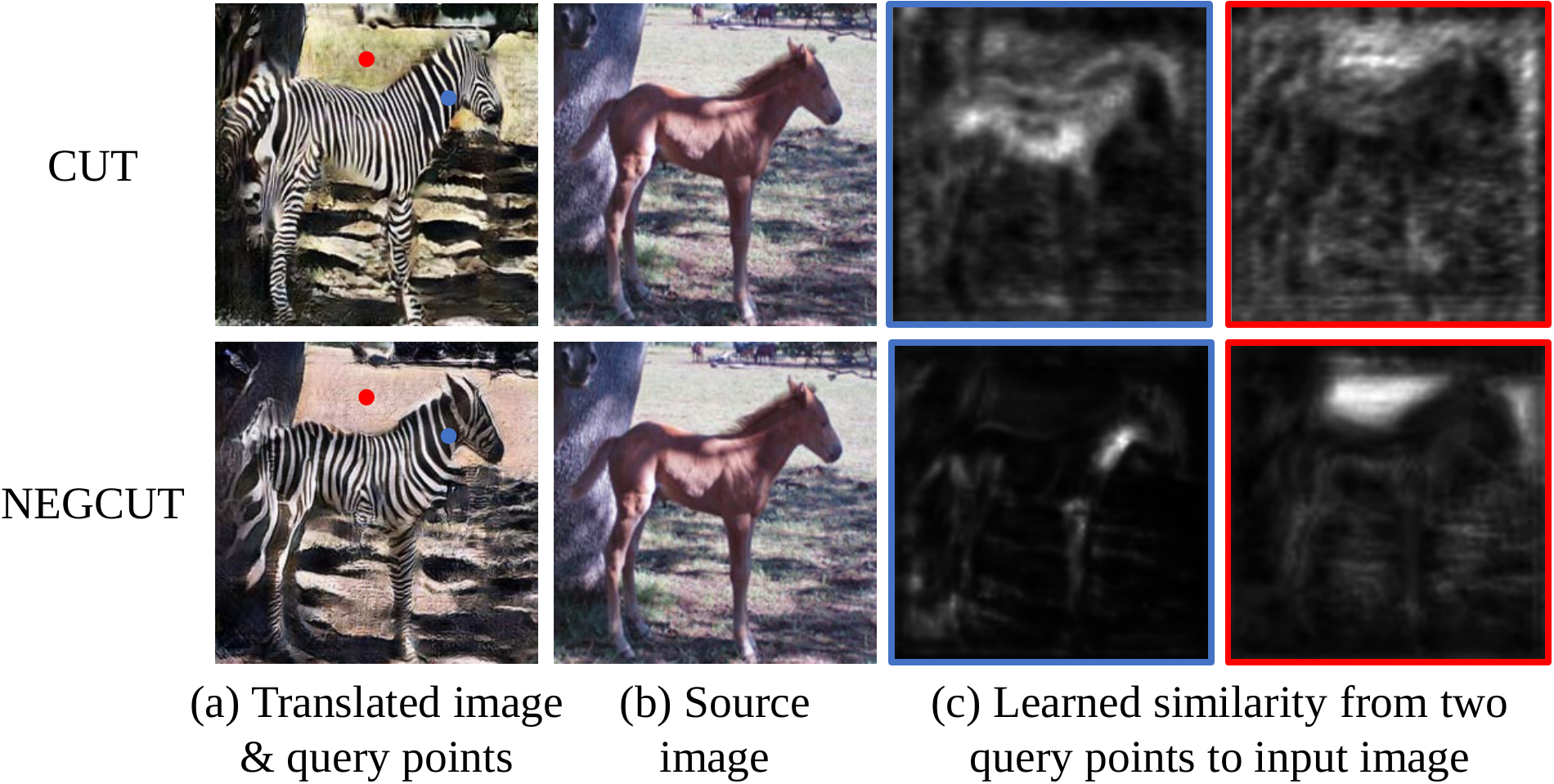}
	\vspace{1mm}
	\caption{Visualization of the learned similarity by representation network in CUT and NEGCUT.
	Two similarity maps are learned from two query points sampled from the foreground~(blue) and background~(red).
	Compared with the similarity learned by CUT, our similarity maps are more concentrated on the neighbourhood of query points, which verifies that our method learns distinguishing representation with the help of hard negative samples.}
	\vspace{-4mm}
	\label{fig:vis}
\end{figure}

Furthermore, we compare the learned similarity by representation network in CUT and NEGCUT.
For each query $q$, we calculate the similarity maps through computing $\mathbf{exp}(\mathbf{q} \cdot \mathbf{k}^+ / \bm{\tau})$ on all the pixels of the image.
From Figure~\ref{fig:vis}, it is observed that, in the similarity maps of CUT, the corresponding areas are scattered over the entire image and several unrelated areas are also associated.
Additionally, when the query point is sampled from a part of the foreground, \emph{i.e.} head of the horse, the whole foreground is associated in the similarity maps of CUT, which demonstrates that the representation network in CUT has difficulty on discriminating different parts of the foreground.
Different from that, the corresponding areas in the similarity maps of NEGCUT are concentrated on the neighborhood of query points or areas with the same semantic, which verifies that the representation network in NEGCUT learns more distinguishing representation and accuracy correspondence under the help of instance-wise hard negative samples.

\section{Conclusion}
\vspace{-0.5em}
In this paper, we propose a novel framework called NEGCUT to mine challenging negative samples for contrastive learning in unpaired image-to-image translation.
Specifically, we design a negative generator trained against the encoder network in an adversarial manner.
The two components in our framework, \emph{i.e.}, the encoder network and the negative generator, are updated alternately to learn distinguishing representation to discriminate positive samples against generated hard negative samples.
Extensive experiments on three benchmark datasets demonstrate the superiority of our method.
Our method achieves state-of-the-art performance and shows a better correspondence between source images and generated images compared with previous methods.

\vspace{-0.5em}

{\footnotesize {\flushleft \bf Acknowledgements}.
This work was supported in part by the National Natural Science Foundation of China under Contract 61836011, 61822208, and 62021001, and in part by the Youth Innovation Promotion Association CAS under Grant 2018497. 
It was also supported by the GPU cluster built by MCC Lab of Information Science and Technology Institution, USTC.
}

{\small
\bibliographystyle{ieee_fullname}
\bibliography{egpaper}
}

\end{document}